# Enhancing Hepatopathy Clinical Trial Efficiency: A Secure, Large Language Model-Powered Pre-Screening Pipeline

Xiongbin Gui[1†], MS, Hanlin Lv[2†], MD, Xiao Wang[2], PhD, Longting Lv[2], MS, Yi Xiao[1, *], MS, Lei Wang[2, 3, *], MS

[1] The First Affiliated Hospital of Guangxi University of Traditional Chinese Medicine, Nanning, 530021, China

[2] Institute of Biointellgence Technology, BGI Research, Wuhan, 430074, China

[3] Guangdong Bigdata Engineering Technology Research Center for Life Sciences, BGI Research, Shenzhen, 518083, China

†These authors contributed equally to this work.

**ABSTRACT**

**Background**: Recruitment for cohorts involving complex liver diseases, such as hepatocellular carcinoma and liver cirrhosis, often requires interpreting semantically complex criteria. Traditional manual screening methods are time-consuming and prone to errors. While AI-powered pre-screening offers potential solutions, challenges remain regarding accuracy, efficiency, and data privacy.

**Methods**: We developed a novel patient pre-screening pipeline that leverages clinical expertise to guide the precise, safe, and efficient application of large language models. The pipeline breaks down complex criteria into a series of composite questions and then employs two strategies to perform semantic question-answering through electronic health records: 1) Pathway A, Anthropomorphized Experts' Chain of Thought strategy; and 2) Pathway B, Preset Stances within an Agent Collaboration strategy, particularly in managing complex clinical reasoning scenarios. The pipeline is evaluated on three key metrics—precision, time consumption, and counterfactual inference—at both the question and criterion levels.

**Results**: Our pipeline achieved high precision (0.921, in criteria level) and efficiency (0.44s per task). Pathway B excelled in complex reasoning, while Pathway A was effective in precise data extraction with faster processing times. Both pathways achieved comparable precision. The pipeline showed promising results in hepatocellular carcinoma (0.878) and cirrhosis trials (0.843).

**Conclusions**: This data-secure and time-efficient pipeline shows high precision in hepatopathy trials, providing promising solutions for streamlining clinical trial workflows. Its efficiency and adaptability make it suitable for improving patient recruitment. And its capability to function in resource-constrained environments further enhances its utility in clinical settings.

**Keywords**

Patient recruitment; Electronic health records; Natural language processing; Liver disease; Eligibility criteria.

# INTRODUCTION

In hepatology, the complexity of chronic diseases like cirrhosis, liver failure, and hepatocellular carcinoma demands meticulous patient selection for clinical research [1]. Clinicians must navigate extensive electronic health records (EHRs) to identify patients with specific diagnoses or treatments [2], such as esophageal varices or those undergoing Transjugular Intrahepatic Portosystemic Shunt (TIPS). The challenge deepens when broader categories like autoimmune diseases or prior beta-blocker therapy are involved, requiring comprehensive evaluation [3]. Additionally, determining whether liver cancer is primary or secondary, identifying the causes of liver failure or cirrhosis, or recognizing symptoms like jaundice adds further layers of complexity, necessitating detailed analysis and clinical insight. These tasks expose the limits of current methods. While traditional manual screening and automated approaches have attempted to address these challenges, they fall short in key areas [4].

Bidirectional Encoder Representations from Transformers (BERT) -based tools have been used for subject matching across various diseases while struggling with generalizability [4-7]. Large language models (LLMs), with their superior contextual semantic understanding, offer a promising avenue to overcome these limitations [8 9]. Domain-specific LLMs, fine-tuned on medical corpora or using knowledge graph embeddings, perform well in clinical text summarization [10 11], note generation [12 13], and medical question answering [14]. However, EHR security policies commonly necessitate local LLM deployment, precluding the use of non-open-source models like Chat Generative Pre-Trained Transformer (ChatGPT), Gemini, or Claude. Additionally, we believe the primary challenge lies in the need for strategic support in complex reasoning rather than in domain-specific knowledge [15]. Thus, both domain-specific and general LLMs start from a similar point. Moreover, domain-specific LLMs' extensive IT resource demands often exceed institutional capacities. Therefore, this study focuses on locally deployed general open-source LLMs for practical reasons.

Open-source LLMs often face criticism in professional reasoning tasks for non-factual hallucination and imbalanced attention focus [16]. To address these issues, effective strategies have been engineered across various domains, including 1) Chain of Thought (CoT) prompting; 2) Agent-Collaboration (Agent-Collab) frameworks.

**CoT Prompting and Extensions:** CoT prompting is a powerful method by breaking down complex tasks into a series of intermediate steps [17]. Traditional "Let's think step by step" approaches with few-shot learning cannot meet our complex reasoning needs. A typical approach involves using prompt engineering techniques to guide the model in exhibiting a thought process consistent with a benchmark individual. In this study, we introduce "Anthropomorphized Experts' CoT" inspired by AI biomimicry. This method emulates human expert reasoning, derived from interviews and summarized into concise steps. The principle is to break down complex thought processes into simple, manageable steps.

**Agent-Collab Frameworks:** Leveraging the capabilities of LLMs in agent mode, where independent agents collaborate, converse, and debate within an Agent-Collab framework, has consistently been of interest. Compared to the consultative question-and-answer model, the Agent-Collab architecture with debating agents excels in all aspects [18]. The Agent-Collab framework typically requires: 1) defining agent roles, and 2) setting the number of dialogue rounds. Roles can be specified or predefined. In this study, we employed a Counterfactual Debating with Preset Stances method, which has shown high consistency in existing research and indicates that two dialogue rounds yield the best results [19]. The Agent-Collab framework can simulate a 'mock trial' scenario to identify and rectify potential errors or biases in interpretation, ensuring a more accurate and comprehensive assessment of patient eligibility. Methods such as self-critique and reflection are also employed to confront the inherent biases of LLMs, compelling them to consider alternative perspectives and generate more accurate outcomes.

To address domain knowledge gaps in locally deployed models, we use powerful non-open-source LLMs to enhance non-sensitive inclusion/exclusion criteria. Building on these advancements, our pipeline transforms complex logical issues into manageable semantic tasks, creating a robust framework for patient pre-screening from admission notes.

# METHODS

## Data preparation

The EHR data used in this study, specifically narrative admission notes, typically included sections such as chief complaint, history of present illness and past medical history (a sample note is provided in the Supplemental Material). These data were obtained from the First Affiliated Hospital of Guangxi University of Chinese Medicine (FAHGUCM). The dataset comprised over 16,000 anonymized admission notes from the Hepatology Department collected over the past 10 years. For detailed analysis, we randomly selected 4,000 records, which was representative of the overall dataset in terms of the distribution of hepatopathy and associated clinical presentations.

We manually reviewed the inclusion and exclusion criteria of six real-world clinical trials on hepatopathy, and carefully selected 58 criteria assessable using information typically documented in admission notes. These criteria primarily focused on general health, medical history, and prior treatments, excluding those requiring specific laboratory results or specialized examinations not routinely captured in admission notes (details in Supplemental Table 1). These trials addressed various liver conditions, including chronic liver failure (ChiCTR2100044187 [20]), cirrhosis (NCT04353193 [21], NCT04850534 [22], NCT03911037 [23], NCT01311167 [24]), and hepatocellular carcinoma (NCT04021056 [25]).

We deployed four LLMs within an intranet security environment: Qwen1.5-7B-Chat-GPTQ-Int4 [26] (QWEN1.5), Baichuan2-13B-Chat [27] (BAICHUAN), ChatGLM3-6B [28] (GLM) and Qwen2-7B-Instruct-GPTQ-Int4 [26] (QWEN2). The experiments were conducted on Lenovo ST650V2 servers equipped with four NVIDIA GeForce RTX 3090 Turbo GPUs.

**Pipeline design**

The core of our approach is a multi-strategy pipeline designed to systematically process narrative admission notes, extracting relevant clinical information and applying logical reasoning. The architecture is detailed in Figure 1.

**Criteria conversion:** We employed an external-CoT strategy, leveraging advanced non-open-source LLMs (OpenAI ChatGPT-4o, Google Gemini Advanced, and Anthropic Claude 3.5 Sonnet) to decompose complex eligibility criteria into simpler and manageable questions. Each LLM generated question sets from a standardized prompt template (see Supplemental Material), which were subsequently integrated and refined using ChatGPT-4o to produce a comprehensive, non-redundant, and conflict/ambiguity-free final question set.

**Question-based assessment:** We employed two distinct pathways to address the decomposed questions.

**Pathway A: Anthropomorphized Experts' CoT**

We conducted interviews with key individuals to summarize their thought processes, which were then used to develop the corresponding CoT. Each LLM, prompted to act in an independent role illustrated in Figure 1(D), analyzed EHR content and provided assessments for corresponding questions: 1) **Clinical Research Coordinator (CRC)**: Focused on identifying relevant EHR domains using empirical knowledge, followed by targeted data extraction. Employed nuanced terminology comprehension beyond basic keyword matching; 2) **Junior Doctor (JD)**: Conducted comprehensive EHR analysis, extracted pertinent information, and generated inferential responses to clinical issues; 3) **Information Engineer (IE)**: Prioritized extracting key terminology from questions, performed global matching within EHR text, and analyzed semantic negations in identified regions. Additionally, we experimented with a majority vote approach, aggregating responses from the three professional perspectives.

**Pathway B: Preset Stance Agent-Collab**

We proposed an Agent-Collab framework simulating a mock trial, wherein three agents with distinct roles and stances independently evaluate questions. To enhance efficiency, the evaluation process is limited to a maximum of two rounds. 1) **Agent A (Positive Assumption) - "Proponent":** Begins with a positive stance (e.g., assuming the patient meets the criteria specified in the question) and conducts a thorough analysis of the admission note to support this assumption, providing a conclusion with supporting evidence; 2) **Agent B (Negative Assumption) - "Opponent":** Adopts a negative stance (e.g., assuming the patient does not meet the criteria in the question) and performs an independent analysis of the same admission note, offering a conclusion with supporting reasoning and evidence; 3) **Agent C (Adjudicator) - "Judge":** Acts as the adjudicator, reviewing the conclusions and reasoning from Agents A and B. If both agents agree, Agent C delivers the consensus conclusion. In the case of disagreement, Agent C identifies inconsistencies and prompts a second, more detailed evaluation round, ultimately providing a final judgement on the question.

**Evaluation metrics**

The pipeline was evaluated on three key metrics—precision, time efficiency, and counterfactual inference rate—at both the question and criterion levels. To establish a gold standard for comparison, clinical experts manually reviewed and annotated the converted question sets from 4,000 admission notes.

We assessed the pipeline's performance at both the question and criterion levels, focusing primarily on precision, while also considering recall, F1 score, and accuracy for a comprehensive evaluation. At the criterion level, we aggregated answers to questions using predefined logical rules (detailed in the Supplemental Material). We conducted an advanced evaluation of the pipeline, including a counterfactual inference analysis [29], by manually reviewing outputs that deviated from the gold standard. Additionally, processing time was measured to verify the pipeline's scalability and efficiency for large-scale pre-screenings. We categorized the questions into relevant clinical domains and further classified them based on reasoning complexity into two types: Classification and Direct Match. This approach allowed us to analyze performance across different clinical domains and task types, providing deeper insight into the pipeline's strengths and limitations.

**Ethical approval and consent to participate**

The study was approved by the Ethics Committee of FAHGUCM (Ref. No. 2020-046-02). The requirement for informed consent was waived by the Ethics Committee due to the observational nature of the study, and all data were de-identified and anonymized.

# RESULTS

We automatically generated 87 questions from 58 original criteria, simplifying complex reasoning into straightforward semantic tasks (Supplemental Table 1). The connection between these criteria and specific trials is detailed in Supplemental Table 2. Various prompt templates were used throughout the pipeline, as outlined in the Supplementary Material.

**Question level assessment**

Pathway B (Preset Stance Agent-Collab) achieved the highest precision at 0.892, outperforming the individual roles in Pathway A (Anthropomorphized Experts' CoTs), which ranged from 0.725 (JD) to 0.827 (CRC). The majority vote strategy in Pathway A further improved its overall precision to 0.877 (Table 1).

**Table 1.** Question level overall performance.

| Pathway/Role | Precision | Recall | F1 | Accuracy |
| --- | --- | --- | --- | --- |
| Pathway A -Role-CRC | 0.827 | 0.804 | 0.780 | 0.978 |
| Pathway A -Role-JD | 0.725 | 0.884 | 0.758 | 0.971 |
| Pathway A -Role-IE | 0.818 | 0.802 | 0.774 | 0.977 |
| Pathway A - Majority Vote | 0.877 | 0.822 | 0.814 | 0.979 |
| Pathway B | 0.892 | 0.793 | 0.809 | 0.972 |

Both pathways demonstrated high annotation consistency, exceeding 97% across all roles and strategies (Figure 2). Additionally, as shown in Figure 3, a substantial proportion of responses achieved precision above 0.85, with 63.2% for Pathway B and 71.3% for Pathway A's majority vote.

To better understand the pipeline's performance, we categorized the derived questions into relevant clinical domains (e.g., Diagnosis, Intervention) and further classified tasks by reasoning complexity (direct match vs. classification). As shown in Table 2, Pathway A excelled in direct match tasks, particularly in the "Diagnosis" (CRC: 0.913) and "Intervention" (CRC: 0.920) categories, highlighting its efficiency in extracting explicit clinical details. Conversely, the "Symptom and Event" category, predominantly composed of classification tasks, posed challenges for all models. Pathway B achieved the highest average precision in this category (0.847), though it performed lower than in other categories, highlighting the complexity of this task type. Role-JD in Pathway A, despite its lower precision, is tailored for tasks requiring broader clinical reasoning, especially where minimizing false negatives is critical (Supplemental Table 3). Pathway B's Agent-Collab mechanism was particularly effective in the "Etiology and Pathology" category (0.921) and in classification tasks across most categories.

**Table 2.** Precision across clinical categories and task types

| Category / Task Type | Pathway A -Role-CRC | Pathway A -Role-JD | Pathway A -Role-IE | Pathway A - Majority Vote | Pathway B |
|---|---|---|---|---|---|
| **Diagnosis** | 0.865 | 0.769 | 0.854 | 0.915 | 0.894 |
| Classification | 0.788 | 0.713 | 0.779 | 0.904 | 0.892 |
| Direct Match | 0.913 | 0.804 | 0.899 | 0.922 | 0.896 |
| **Etiology and Pathology** | 0.797 | 0.659 | 0.789 | 0.869 | 0.910 |
| Classification | 0.759 | 0.594 | 0.737 | 0.876 | 0.921 |
| Direct Match | 0.853 | 0.756 | 0.867 | 0.859 | 0.895 |
| **Symptom and Event** | 0.637 | 0.512 | 0.635 | 0.726 | 0.847 |
| Classification | 0.637 | 0.512 | 0.635 | 0.726 | 0.847 |
| **Intervention** | 0.812 | 0.721 | 0.806 | 0.837 | 0.892 |
| Classification | 0.740 | 0.627 | 0.740 | 0.774 | 0.880 |
| Direct Match | 0.920 | 0.862 | 0.905 | 0.933 | 0.909 |
| **Overall Average** | 0.827 | 0.725 | 0.818 | 0.877 | 0.892 |

**Criterion level performance**

At the criterion level, precision in Pathway A ranged from 0.787 for JD to 0.893 for CRC, with the majority vote boosting it to 0.922. Pathway B achieved a precision of 0.920, comparable to Pathway A's majority vote (Table 3).

**Table 3.** Criterion level overall performance.

| Pathway/Role | Precision | Recall | F1 | Accuracy |
|---|---|---|---|---|
| Pathway A -Role-CRC | 0.893 | 0.808 | 0.818 | 0.974 |
| Pathway A -Role-JD | 0.787 | 0.868 | 0.807 | 0.965 |
| Pathway A -Role-IE | 0.887 | 0.803 | 0.814 | 0.972 |
| Pathway A - Majority Vote | 0.922 | 0.819 | 0.835 | 0.975 |
| Pathway B | 0.920 | 0.785 | 0.820 | 0.964 |

**Counterfactual inference**

We evaluated counterfactual inference rates across the pathways. Pathway B had the lowest rate at 0.25%, indicating strong consistency. In Pathway A, the rates varied among roles, with CRC at 0.82%, IE at 0.93%, and JD at 1.78%. The majority vote approach effectively reduced Pathway A's overall counterfactual inference rate to 0.77%, demonstrating the benefit of collective decision-making in minimizing errors.

**Time consumption**

To evaluate efficiency, we measured the time spent on question assessment using an optimal concurrency level of 3 to balance speed and resource use. Pathway A showed significantly faster processing times, averaging 0.588 seconds per question for CRC, 0.395 seconds for JD, and 0.340 seconds for IE. In contrast, Pathway B took an average of 2.570 seconds per question (Figure 4).

Supplementary Figure 1 illustrates the variability in processing times for both pathways across different questions. Pathway B exhibited a wider range due to the possibility of a second evaluation round. Overall, our pipeline demonstrated efficient and manageable processing times. With further optimization, Pathway A could potentially process a single patient's eligibility in approximately 10 seconds.

**Additional research**

We also conducted a preliminary evaluation of other locally deployed LLMs (QWEN1.5, BAICHUAN, and GLM) on a subset of the data. QWEN1.5 generally outperformed the other models. BAICHUAN showed a tendency for over-inference, while GLM faced challenges with constraint adherence.

Furthermore, we compared the performance of QWEN1.5 and QWEN2 when deployed locally. Interestingly, QWEN1.5 outperformed QWEN2, despite the latter's enhancements in reasoning capabilities. This discrepancy might be attributed to QWEN2's conservative approach, leading to a higher rate of false negatives, or the lack of prompt adaptation for QWEN2 in our specific setting.

## DISCUSSION

**Principle Findings**

In this study, the pipeline combined anthropomorphized experts' CoT with a preset stance Agent-Collab strategy and a locally deployed LLM, effectively addressing the challenges in clinical trial pre-screening for hepatology by achieving acceptable precision and improving efficiency. Notably, the entire pipeline operated on a single consumer-grade server within a fully air-gapped data, internet-isolated environment.

In conclusion, we propose a hybrid approach for clinical application: 1) For non-inferential recognition tasks involving explicit diagnoses and interventions, pathway A with majority vote is recommend; 2) For types requiring complex reasoning, Pathway B is more suitable.

**Outperforming across multiple diseases in secure environments**

The pipeline achieved a competitive average precision of 0.921 at the criterion level, which was achieved within a secure environment with limited resources. Additionally, we assessed the trial-level precision, achieving an average precision of 0.72, considering the limitations of admission notes in determining patient eligibility. Despite these constraints, the pipeline achieved higher precision in hepatocellular carcinoma (0.878) and cirrhosis trials (0.843), indicating its potential utility in hepatopathy trials. Compared to similar studies in Table 4, our pipeline outperformed, even when using a closed-source LLM.

**Table 4.** Performance comparison of related research.

| Research | Patient | Model | Strategy | Precision |
|---|---|---|---|---|
| Thai et al., 2024 [15] | 1,400 | ChatGPT 4 | Automatic Cohort Retrieval | 0.794 |
| Jin et al., 2024 [30] | 184 | ChatGPT 3.5 | TrialGPT | 0.676 |

| Nievas et al., 2023 [31] | 4,678 | ChatGPT 3.5 | CoT predicting | 0.603 |
| | | ChatGPT 4 | | 0.701 |
| | | LLAMA | | 0.396 |
| Kusa et al., 2023 [32] | 125 | BERT | Supervised learning | 0.291 |

For tasks such as identifying hepatocellular carcinoma, cirrhosis, or specific interventions like TIPS, the CoT-based approach delivered high precision (a typical of 0.92, verse traditional manual annotation methods which achieve around 0.81, reflecting a 14% improvement) with minimal processing time (a typical of 0.44 seconds per task, verse traditional manual annotation which take over 10 seconds per task, representing a 95% reduction) [33]. For more complex tasks, such as determining the etiology of liver failure, assessing whether liver cancer is primary, identifying systemic diseases, or verifying HBV antiviral treatment, the Agent-Collab framework provided even higher precision. By leveraging externally defined logic—derived from advanced LLMs or clinical experts—this combination ensured that our method was both efficient and adaptable, effectively handling a wide range of clinical scenarios in hepatopathy research.

**Clinical expertise guides LLM application to tackle complex challenges**

Researchers often view new technologies and standard clinical practices separately. In this study, we integrated proven clinical methods with advanced technologies to effectively address complex problems, yielding promising results.

Pathway A's anthropomorphized experts guided the LLM from general semantic understanding to targeted entity recognition, a critical feature for enhancing the performance of weaker open-source LLMs. Meanwhile, Pathway B's preset stance Agent-Collab strategy effectively managed complex and semantically disordered texts, ensuring accurate outputs. By combining advanced LLM capabilities with clinical expertise, both strategies showed potential for improving the accuracy of complex clinical assessments (Figure 5).

While Pathway B showed marked improvements in complex reasoning and nuanced categorization, Pathway A excelled in tasks involving direct recognition of explicit terms, with its CRC and IE roles demonstrating strong proficiency in precise data extraction for direct match tasks. This contrast highlights their complementary strengths. Although Pathway B is generally more robust, Pathway A offers comparable accuracy with significantly lower resource consumption, making it a practical choice when efficiency is paramount.

**Exploratory insights on pipeline adaptability in traditional Chinese medicine**

Notably, our dataset, which comes from FAHGUCM in China, a regionally influential hospital in hepatopathy (annual outpatient visits reach 50,000, with around 5,000 inpatient admissions). This dataset integrates both Traditional Chinese Medicine (TCM) and Western medicine, included expressions influenced by TCM. While not being the primary focus of our evaluation, these expressions were also tested, and preliminary results suggested that the pipeline can adapt to these diverse medical terminologies. Some natural language expressions in the recorded texts reflect a TCM influence. Although these expressions were not involved in the nano-ranking Q&A, we experimented with incorporating some TCM phrasing in our tests. For example, we used criterion like "whether the patient has insomnia", and the corpus was "患者寐差" (poor sleep quality) or "卧不安寐" (lie unable to sleep) instead of the more commonly seen "患者存在失眠情况" (the patient has insomnia condition). The recognition performance was satisfactory, suggesting potential applicability in this area. However, we did not conduct systematic and rigorous testing on these variations.

**Limitations**

Our study's primary limitation lies in the restricted data dimensions within admission notes, which focus on key sections like chief complaints, current disease history, and past medical history. To address this, we selected criteria from multiple clinical trials related to hepatopathy and manually filtered them to ensure relevance to our dataset. The study was conducted using data from a single center and it requires further external validation to ensure its generalizability.

Despite these limitations, our approach provides valuable insights into the use of limited EHR data for clinical trial matching. By focusing on likely available data points, we enhance the real-world applicability of our analysis. This focused methodology also facilitates scalability to other settings with similar data limitations. Future work should enhance data collection in EHR systems, improve model robustness for incomplete data, and expand this approach to other medical domains.

**CONCLUSION**

We have developed an effective patient pre-screening strategy, showing high precision in hepatopathy trials. Designed to operate efficiently in environments with limited IT resources and stringent security requirements, our pipeline combines different approaches to achieve both precision and efficiency across various clinical tasks. This cost-effective and adaptable solution has the potential to enhance the pre-screening process and improve patient recruitment in clinical trials.

**Data Availability Statement**

The data analyzed in this study are not publicly available due to privacy or ethical restrictions but can be obtained from the corresponding author upon reasonable request.

**Financial Support and Sponsorship**

This study was supported by the Second Batch of National TCM Clinical Research Base Construction Units (No. 131 [2018], issued by the National Administration of Traditional Chinese Medicine).

**Conflict of Interests**

The authors report no conflict of interest.

**Ethics Approval Statement**

The study was approved by the First Affiliated Hospital of Guangxi University of Chinese Medicine (approved Ref. No. 2020-046-02).

**Patient Consent Statement**

The requirement for informed consent was waived by the Ethics Committee of the First Affiliated Hospital of Guangxi University of Chinese Medicine, due to the retrospective nature of the study, and all clinical data were de-identified and anonymized.

**Author Contributions**

Xiongbin Gui: Conceptualization, Methodology, Writing - Original Draft, Formal Analysis

Hanlin Lv: Methodology, Formal Analysis, Software, Writing - Review & Editing

Xiao Wang: Data Curation, Validation, Visualization, Supervision

Longting Lv: Investigation, Project Administration, Writing - Review & Editing

Yi Xiao: Conceptualization, Funding Acquisition, Writing - Review & Editing, Supervision

Lei Wang: Methodology, Project Administration, Writing - Review & Editing, Supervision

**Acknowledgment**

We sincerely thank the China National GeneBank for their technical support. We also appreciate the support from RUIYI's Clinical Multi-omics Data Research Workstation for our research.

7. Tissot HC, Shah AD, Brealey D, et al. Natural Language Processing for Mimicking Clinical Trial Recruitment in Critical Care: A Semi-Automated Simulation Based on the LeoPARDS Trial. IEEE J Biomed Health Inform 2020;**24**(10):2950-59 doi: 10.1109/JBHI.2020.2977925 [published Online First: 2020/03/10].

8. Lee P, Bubeck S, Petro JJNEJoM. Benefits, limits, and risks of GPT-4 as an AI chatbot for medicine. 2023;**388**(13):1233-39.

9. Lee P, Goldberg C, Kohane I. *The AI revolution in medicine: GPT-4 and beyond*: Pearson, 2023.

10. Van Veen D, Van Uden C, Blankemeier L, et al. Adapted large language models can outperform medical experts in clinical text summarization. Nat Med 2024;**30**(4):1134-42 doi: 10.1038/s41591-024-02855-5 [published Online First: 2024/02/28].

11. Devarakonda MV, Mohanty S, Sunkishala RR, Mallampalli N, Liu XJJobi. Clinical trial recommendations using Semantics-Based inductive inference and knowledge graph embeddings. 2024;**154**:104627.

12. Van Veen D, Van Uden C, Blankemeier L, et al. Clinical text summarization: Adapting large language models can outperform human experts. 2023.

13. Yuan D, Rastogi E, Naik G, et al. A Continued Pretrained LLM Approach for Automatic Medical Note Generation. 2024.

14. Singhal K, Tu T, Gottweis J, et al. Towards Expert-Level Medical Question Answering with Large Language Models. 2023. https://ui.adsabs.harvard.edu/abs/2023arXiv230509617S (accessed May 01, 2023).

15. Thai DN, Ardulov V, Mena JU, et al. ACR: A Benchmark for Automatic Cohort Retrieval. 2024.
**22** / **30**

**Figures, Figure Captions and Legends**

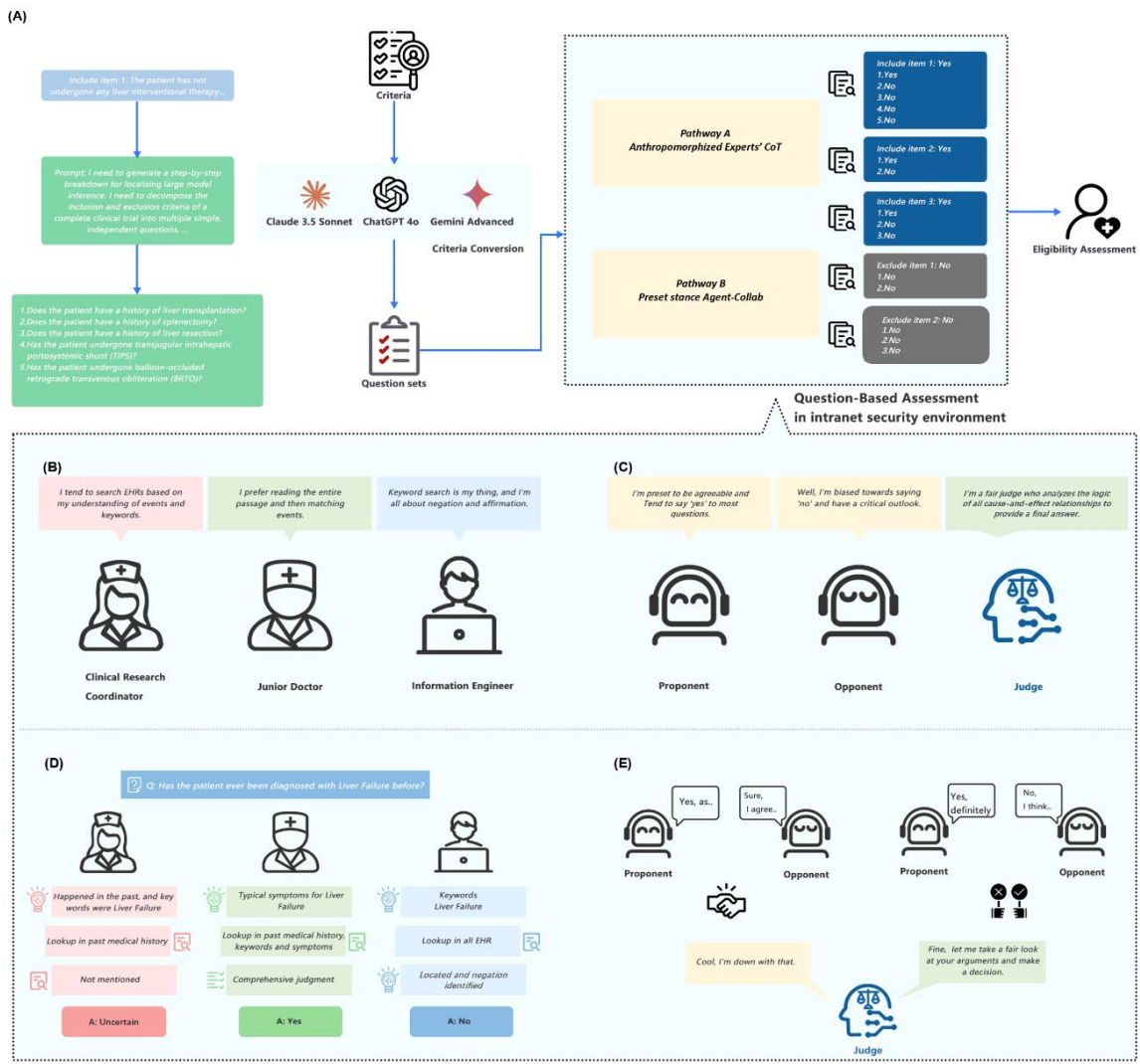

**Fig. 1 Comparison of two pathways for criteria identification**

(A) shows the complete pre-screening process, including criteria conversion and question-based assessment. (B) and (D) depict Pathway A: Anthropomorphized Experts' Chain of Thought (CoT), which models the opinions of three real-world operators to create a CoT that aligns with inclusion and exclusion criteria. (C) and (E) illustrate Pathway B: Preset Stance Agent-Collab, where agents' stances are preset, followed by a voting and arbitration process to determine final answers.

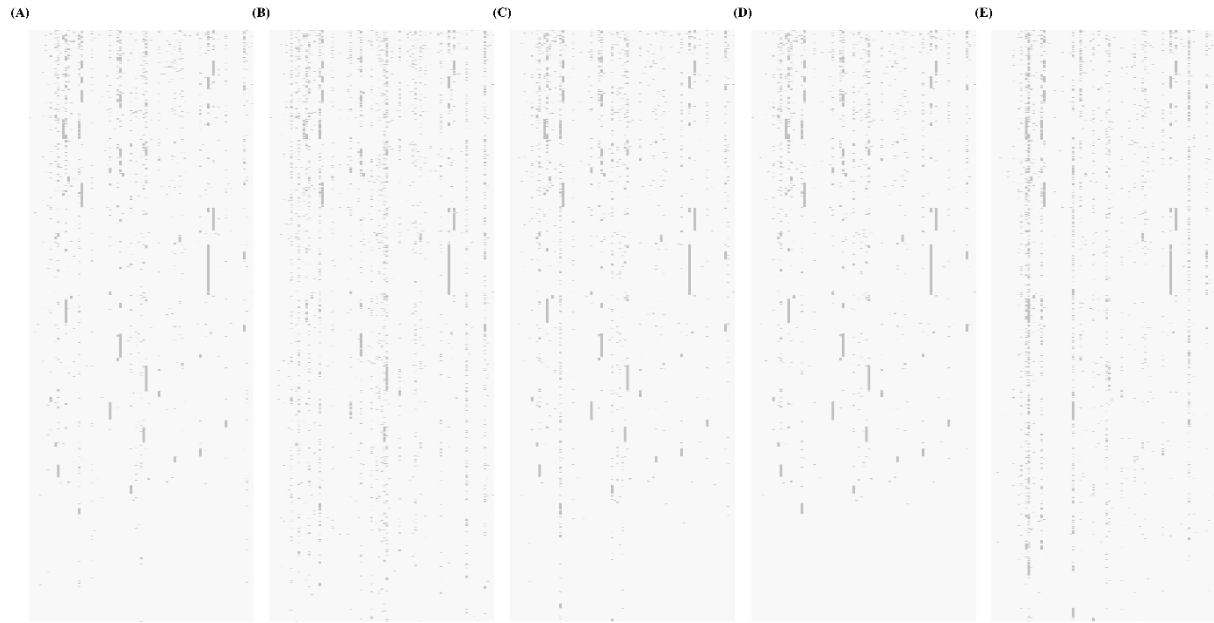

**Fig. 2 Heatmap of annotation consistency across admission records**

Rows correspond to questions and columns to admission records. Each cell indicates whether the answer is consistent (white) or inconsistent (gray) based on expert annotations. Panels (A) through (E) show results for different pathways: (A) Pathway A-Role-CRC, (B) Pathway A-Role-JD, (C) Pathway A- Role-IE, (D) Pathway A-majority vote, and (E) Pathway B.

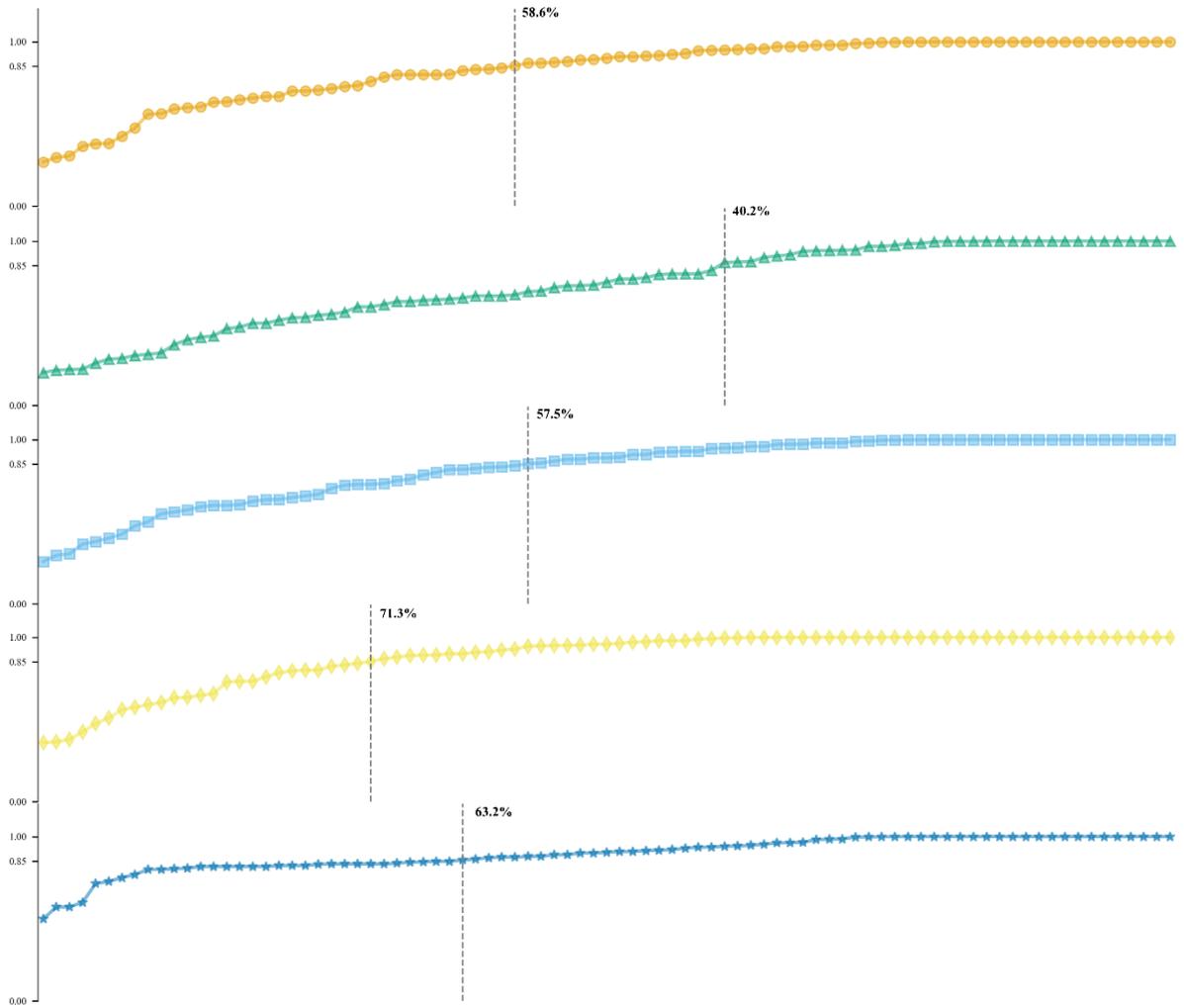

**Fig. 3 Precision across questions for varying pathways**

This figure shows question-level precision for each strategy: (A) Pathway A- Role-CRC, (B) Pathway A- Role-JD, (C) Pathway A- Role-IE, (D) Pathway A-majority vote and (E) Pathway B. Questions are ordered by increasing precision along the X-axis, with precision values on the Y-axis. Vertical dashed lines indicate questions with a precision of 0.85.

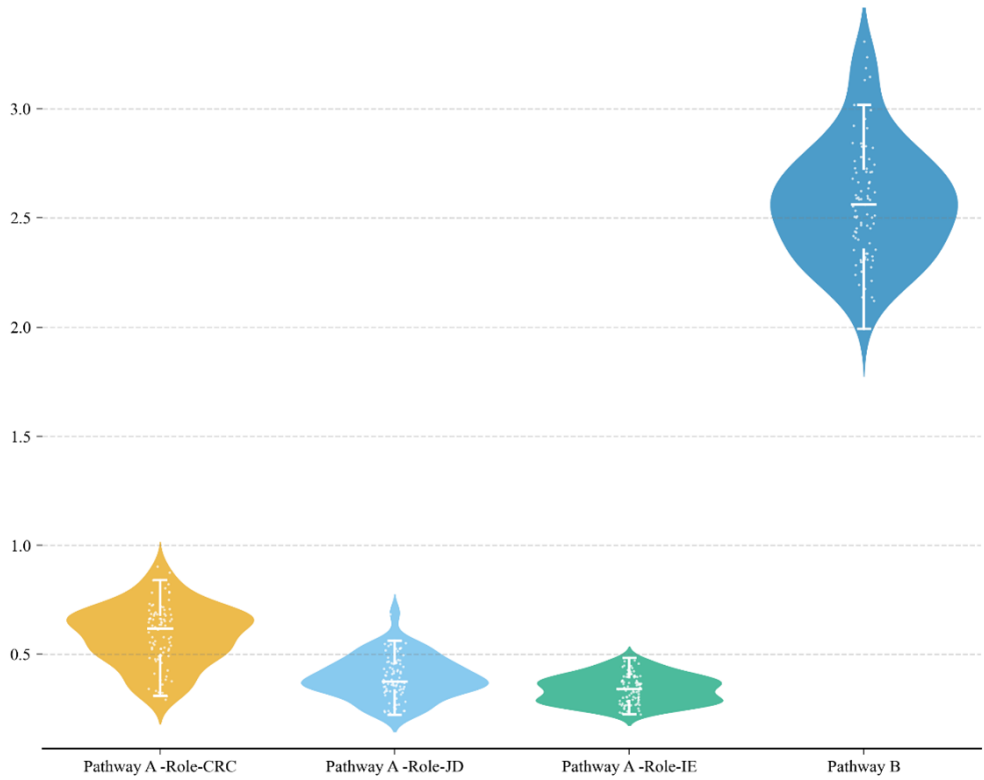

**Fig. 4 Processing time across pathways**

This figure illustrates the distribution of processing times for different strategies: Pathway A-Role-CRC, Pathway A- Role-JD, Pathway A- Role-IE, and Pathway B.

| Step | Content | Answer |
|---|---|---|
| **Original Criterion** | Has the patient been diagnosed with primary liver cancer? | "Unable to determine". The text mentions a diagnosis of liver cancer but does not specify whether it is primary liver cancer. |
| **Converted Questions** | Q1. Has the patient been diagnosed with a malignant liver tumor? | "Yes". The patient was diagnosed with liver cancer. which is a malignant liver tumor. |
| | Q2. Is the pathological type of the liver tumor hepatocellular carcinoma? | "Yes". The pathological type is specified as hepatocellular carcinoma (trabecular type). |
| | Q3. Has the patient been diagnosed with mixed hepatocellular carcinoma? | "No". The pathological type is specified as hepatocellular carcinoma (trabecular type). |
| | Q4. Is there any mention that the liver tumor metastasized from another site? | "Information not provided". There is no indication that the tumor metastasized form another site. |
| **Final Determination** | The original question is determined to be "Yes" if:<br>• Q1 is answered "Yes"<br>• AND any of Q2 or Q3 is answered "Yes"<br>• AND Q4 is not answered "Yes" | "Yes". |

*Admission note*

History of liver cancer resection over 9 years ago, with two recurrences in the past 2 years. The patient presented to our hospital in April 2009 due to 'liver mass detected on examination.' CT revealed a mass in the lower segment of the posterior right lobe of the liver. On June 29, 2009, the patient underwent right hepatectomy for liver cancer. Postoperative pathology indicated hepatocellular carcinoma (trabecular type).

**Fig. 5 Decomposition and answer example for complex questions**

This figure demonstrates how decomposing the complex question "Has the patient been diagnosed with primary liver cancer?" into four sub-questions improves model performance, leading to a final "Yes" determination based on rule-based criteria.

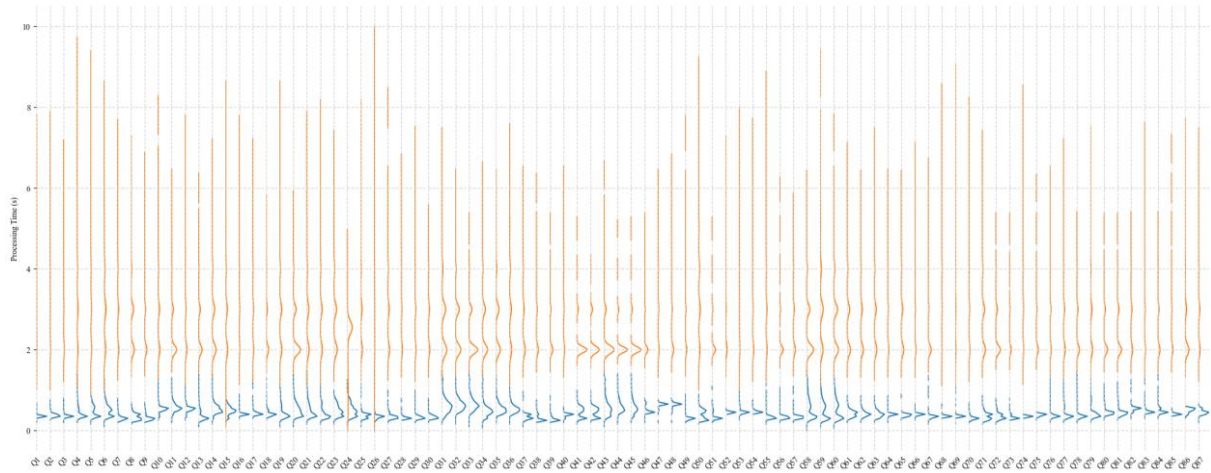

**Supplemental Fig. 1 Processing time across questions**

Distribution of processing times for Pathway A and Pathway B across questions. Pathway A combines similar processing times across three roles, while Pathway B includes total time for one or two rounds per case. The X-axis represents questions, and the Y-axis shows processing time per case. Blue and yellow lines represent Pathway A and Pathway B, respectively.